\documentclass[conference]{IEEEtran}
\IEEEoverridecommandlockouts
\usepackage{graphicx}
\usepackage{cite}
\usepackage{amsmath,amssymb,amsfonts}
\usepackage{algorithmic}
\usepackage{graphicx}
\usepackage{textcomp}
\usepackage{hyperref}
\hypersetup{
    colorlinks=true,
    linkcolor=black,  
    citecolor=black,   
    urlcolor=black
}
\usepackage{xcolor}
\def\BibTeX{{\rm B\kern-.05em{\sc i\kern-.025em b}\kern-.08em
    T\kern-.1667em\lower.7ex\hbox{E}\kern-.125emX}}
\begin{document}

\title{NutrifyAI: An AI-Powered System for Real-Time Food Detection, Nutritional Analysis, and Personalized Meal Recommendations\\
}

\author{
    \IEEEauthorblockN{Michelle Han}
    \IEEEauthorblockA{
        \textit{New York University Stern} \\
        New York, New York \\
        mh7793@nyu.edu
    }
    \and
    \IEEEauthorblockN{Junyao Chen}
    \IEEEauthorblockA{
        \textit{New York University Stern} \\
        New York, New York \\
        junyao.chen@nyu.edu
    }
    \and
    \IEEEauthorblockN{Zhengyuan Zhou}
    \IEEEauthorblockA{
        \textit{New York University Stern} \\
        New York, New York \\
        zzhou@stern.nyu.edu
    }
}
\maketitle
\begin{abstract}
With diet and nutrition apps reaching 1.4 billion users in 2022 [1], it’s no surprise that popular health apps, MyFitnessPal, Noom, and Calorie Counter, are surging in popularity. However, one major setback [2] of nearly all nutrition applications is that users must enter food data manually, which is time-consuming and tedious. Thus, there has been an increasing demand for applications that can accurately identify food items, analyze their nutritional content, and offer dietary recommendations in real-time. This paper introduces a comprehensive system that combines advanced computer vision techniques with nutrition analysis, implemented in a versatile mobile and web application. The system is divided into three key components: 1) food detection using the YOLOv8 model, 2) nutrient analysis via the Edamam Nutrition Analysis API, and 3) personalized meal recommendations using the Edamam Meal Planning and Recipe Search APIs. Preliminary results showcase the system’s effectiveness  by providing immediate, accurate dietary insights, with a demonstrated food recognition accuracy of nearly 80\%, making it a valuable tool for users to make informed dietary decisions.
\end{abstract}

\begin{IEEEkeywords}
computer vision, YOLO model, artificial intelligence, machine learning, food recognition

\end{IEEEkeywords}

\section{Introduction}
The rise of artificial intelligence has significantly impacted various aspects of human daily life, ranging from healthcare to entertainment. Among the numerous advancements in artificial intelligence, computer vision stands out for its potential in how users interact and interpret visual data. Particularly, computer vision offers a promising application in the domain of food recognition and nutrient analysis [3]. This research aims to explore the capabilities of a computer vision model, specifically a YOLO (You Only Look Once) model, to accurately recognize food dishes and perform a reliable analysis of their nutritional content. YOLO models are a family of real-time object detection algorithms that excel at processing images in a single evaluation, predicting bounding boxes and class probabilities simultaneously. Unlike traditional object detection models, which process images in multiple stages, YOLO’s single-stage approach allows for much faster detection times without compromising accuracy [4]. This makes YOLO models particularly suitable for this application, which requires rapid and precise object detection. In this paper, variations of the YOLOv5 and YOLOv8 models are tested.

\section{Literature Review} Table~\ref{table:lit_review} reviews key studies relevant to the development of NutrifyAI, focusing on food detection using YOLO models, the integration of nutritional APIs for real-time analysis, and AI-driven meal recommendations.

\begin{table*}[htbp]
\caption{Summary of Related Work in Food Detection and Nutritional Analysis}
\begin{center}
\begin{tabular}{|p{2.7cm}|p{2cm}|p{3.7cm}|p{3.8cm}|p{3.8cm}|}
\hline
\textbf{Title} & \textbf{Authors} & \textbf{Description} & \textbf{Benefits} & \textbf{Limitations} \\
\hline
FoodTracker: A Real-time Food Detection Mobile Application by Deep Convolutional Neural Networks [8]
& Jianing Sun, Katarzyna Radecka, Zeljko Zilic
& Describes a mobile application that recognizes multi-object meals from a single image and provides nutritional facts, utilizing YOLOv2 and deep convolutional networks.
& Efficient and fast food detection suitable for mobile devices; provides real-time nutritional analysis which enhances user convenience and encourages consistent use.
& Relies on a pre-trained DCNN that may not generalize well to foods outside the training dataset, which was primarily composed of Japanese food items. Limited by mobile device computational resources. \\
\hline
Enhanced Food Classification System Using YOLO Models For Object Detection Algorithm [9]
& Sudharson S, Priyanka Kokil, Annamalai R, N V Sai Manoj
& The paper evaluates the effectiveness of YOLOv5 and YOLOv7 models in food classification tasks using a custom dataset.
& Comprehensive results: mAP, recall, precision, F1; YOLOv5 showed superior performance in accuracy, recall, and mAP. Suitable for real-time applications in food recognition.
& Uses a relatively small dataset of 150 photos; only compared 5 foods (4 Asian, 1 American). \\
\hline
A Framework to Identify Allergen and Nutrient Content in Fruits and Packaged Food Using Deep Learning and OCR [10]
& B Rohini, Divya Pavuluri, LS Kumar, V Soorya, Aravinth Jagatheesan
& Combines OCR with deep learning for identifying allergens and nutrients on food packaging.
& Enhances consumer safety by providing detailed information about food products.
& Limited to packaged foods; may not generalize to loose or natural foods. \\
\hline
Evaluating the Effectiveness of YOLO Models in Different Sized Object Detection and Feature [11]
& Luyl-Da Quach, Khang Nguyen Quoc, Anh Nguyen Quynh, and Hoang Tran Ngoc
& Assesses YOLO models' performance in classifying small objects, highlighting challenges with background noise and feature loss.
& Demonstrates the capability of YOLO models in handling small object detection, which is pertinent for accurately identifying small or intricately shaped food items.
& The focus on small objects may not translate directly to the variety and complexity of food items typically encountered in broader dietary tracking applications. \\
\hline
Real-Time Flying Object Detection with YOLOv8 [12]
& Reis, D., Hong, J., Kupec, J., Daoudi, A.
& This paper presents a generalized and refined model for real-time detection of flying objects, utilizing YOLOv8 for high precision and inference speed.
& Offers insights into optimizing object detection systems for challenging conditions like high-speed and variable-sized targets, useful for drone detection and other applications.
& Focuses primarily on flying objects, which might limit direct applicability to other types of real-time object detection scenarios. \\
\hline
\end{tabular}
\label{table:lit_review}
\end{center}
\end{table*}

\section{Methodology}
\subsection{Overview}
This research explores the potential of AI technologies in enhancing daily food tracking through detection, analysis, and recommendations. NutrifyAI, a web application, was developed to highlight its practicality and usability in everyday life. By offering a user-friendly interface, the app allows individuals to seamlessly integrate nutritional tracking into their routines. Users can simply capture an image of their meal, which the app processes to detect and identify the food items present.

Leveraging the power of the YOLOv8 computer vision model, NutrifyAI accurately recognizes various food items within the image. Once detected, the app retrieves detailed nutritional information for each identified food item using the Edamam API. This data is then aggregated to provide users with a comprehensive nutritional breakdown, including key metrics such as total calories, fat, protein, carbohydrate, and fiber content.

Beyond tracking, NutrifyAI builds a history of the foods scanned by the user. Using this data, the app offers personalized meal recommendations tailored to the user's nutritional goals.

\begin{figure}[h]
    \centering
    \includegraphics[width=\linewidth]{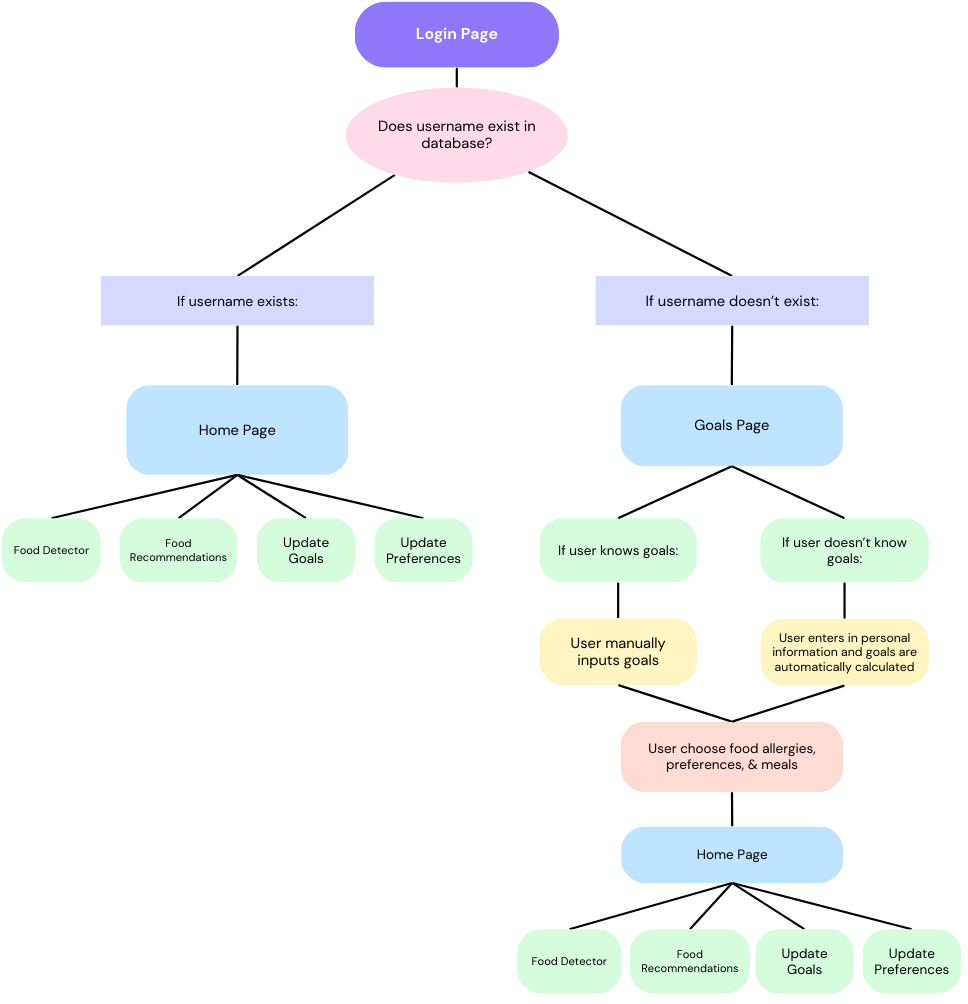}
    \caption{End-to-end User Pipeline for NutrifyAI}
    \label{fig:app pipelline}
\end{figure}

\subsection{Data Collection and Preprocessing}
\subsubsection{Training Datasets}
The training of the model was based on datasets that had already been aggregated and preprocessed by the repository available at GitHub [5]. 5 datasets were used:

\paragraph{Open Images V6-Food Dataset} Over 20,000 food-related images were extracted from Google's Open Images V6, focusing on 18 specific food labels of American cuisine. 

\paragraph{School Lunch Dataset} This dataset includes 3,940 images of Japanese high school lunches, categorized into 21 distinct classes plus an "Other Foods" category. It was used to introduce real-world variability in food presentation.

\paragraph{Vietnamese Food Dataset} Images of 10 traditional Vietnamese dishes (e.g., Pho, Com Tam) were added, with about 20-30 images per dish, to enhance cultural diversity in the model's training.

\paragraph{MAFood-121 Dataset} Containing 21,175 images of 121 dishes from the world's top 11 cuisines, this dataset helps the model learn a broad range of global foods, with 85\% used for training.

\paragraph{Food-101 Dataset} This dataset includes 101,000 images across 101 dish types, with 750 training and 250 testing images per dish, contributing to a comprehensive training environment.

The final expanded dataset comprised 93,748 training images and 26,825 evaluation images, covering a total of 180 distinct dishes. 

\subsection{Model Architecture}
The YOLOv8 model was trained using the aforementioned datasets, each labeled with food categories and bounding boxes. The dataset was split into 80\% for training and 20\% for validation. To enhance the model's generalization ability, data augmentation techniques were applied such as rotations, scaling, and flips to simulate different viewing conditions.

The training process used Stochastic Gradient Descent with momentum. Starting with an initial learning rate of 0.01, adjustments were made via a learning rate scheduler to ensure consistent improvement and prevent overfitting. The model was trained for 50 epochs, with early stopping used to halt training if the validation loss did not improve after 10 consecutive epochs.

In addition to YOLOv8, EfficientNet-B4 was used to fine-tune on the dataset, starting with pre-trained weights from ImageNet. The fine-tuning process involved training for 30 epochs using the Adam optimizer with a learning rate of 0.001, with similar data augmentation techniques applied to improve the model's robustness in recognizing diverse food items.

\subsection{Model Evaluation}
The evaluation of the model was conducted using the Food Recognition 2022 dataset [6], a comprehensive annotated dataset specifically designed for semantic segmentation tasks, which includes 43,962 images with 95,009 labeled objects belonging to 498 different classes. The GitHub repository [5] from which the training data was obtained tested five variations of the YOLO model with results measured based on mAP at 0.5 IoU. The chart in Fig. 2 presents the mAP results for each YOLO variant, highlighting the superior performance of YOLOv8s with a mAP of 0.963. This high performance is why YOLOv8s powers the NutrifyAI application. 

\begin{figure}[h]
    \centering
    \includegraphics[width=\linewidth]{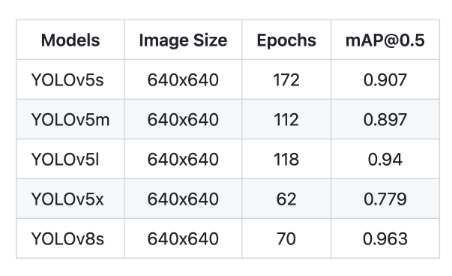}
    \caption{Comparison of mAP results across YOLO model variations at 0.5 IoU [5]}
    \label{fig:chart}
\end{figure}

To assess the model’s performance, a subset of the Food Recognition 2022 dataset was selected that intersected with the classes present in the training data. This resulted in a test set comprising 55 food categories. Given that the Food Recognition 2022 dataset contains 498 different classes, many of which were not present in the training data, a crucial step in the evaluation process was label mapping. This involved aligning the labels from the Food Recognition 2022 dataset with those used during training. For example, while the original training dataset had a single class labeled as "wine," the Food Recognition 2022 dataset included more specific classes such as "rose-wine" and "red-wine." These specific classes were mapped back to the more general "wine" label to ensure consistency in evaluation.

\subsection{API Integration}
\paragraph{Edamam Nutrient Analysis API and Google Sheets API} This API retrieves detailed nutritional data (calories, fat, protein, etc.) for food items detected by the YOLOv8 model. After the request is sent to the API, the data is fetched and visualized using Chart.js (Fig. 4) on the user interface and is stored in a Google Sheets database (Fig. 3, Fig. 6) for tracking and further analysis. 

\begin{figure}[h]
    \centering
    \includegraphics[width=\linewidth]{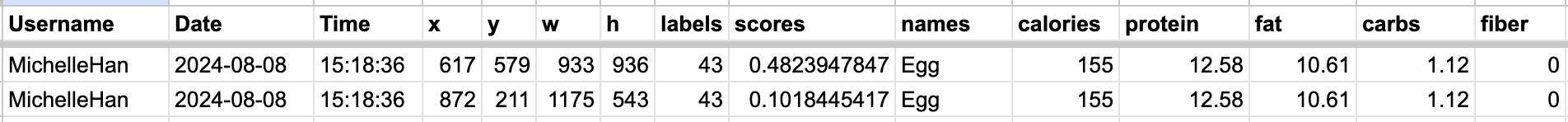}
    \caption{Food tracking example in Google Sheets}
    \label{fig:app gsheet2}
\end{figure}

\begin{figure}[h]
    \centering
    \includegraphics[width=\linewidth]{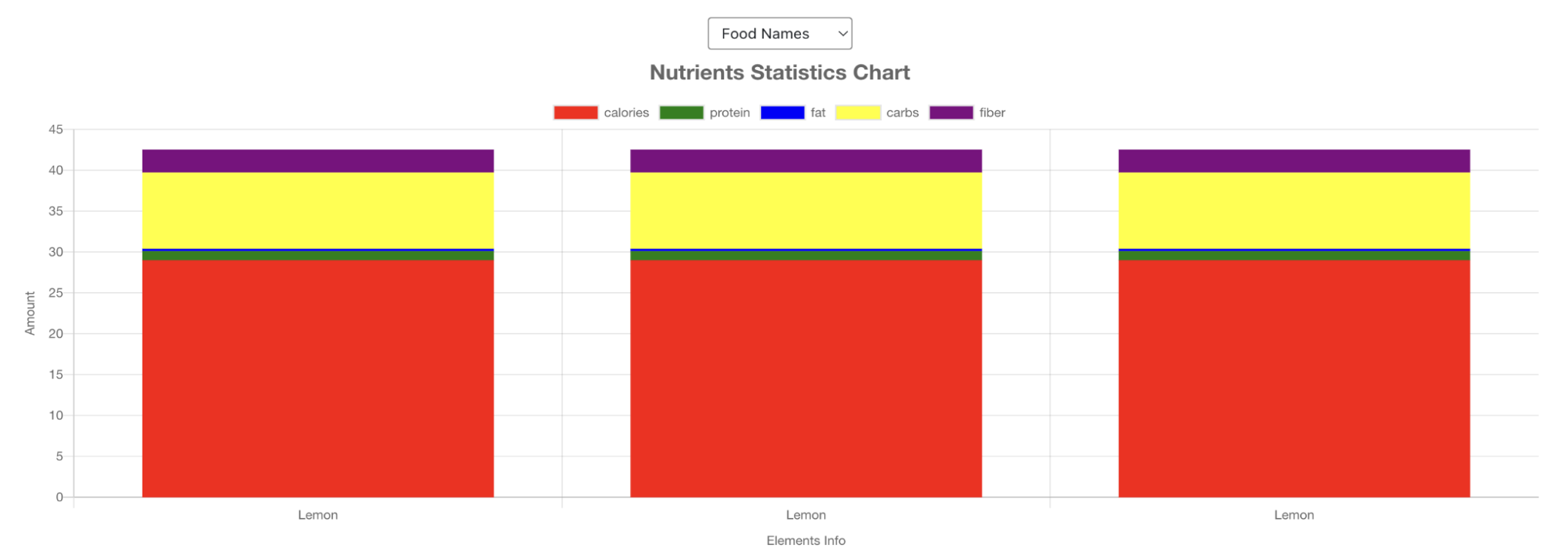}
    \caption{Nutrient analysis chart on web-app interface}
    \label{fig:chart.js}
\end{figure}

\paragraph{Edamam Recipe and Meal Planning API} These APIs provide personalized meal recommendations based on the user's nutritional goals, suggesting recipes and meal plans that align with the user's dietary preferences, restrictions, and nutritional targets. Recommendations are generated based on entered or calculated nutritional goals (Fig. 5), offering tailored meal suggestions to help users meet their health objectives. 
\begin{figure}[h]
    \centering
    \includegraphics[width=\linewidth]{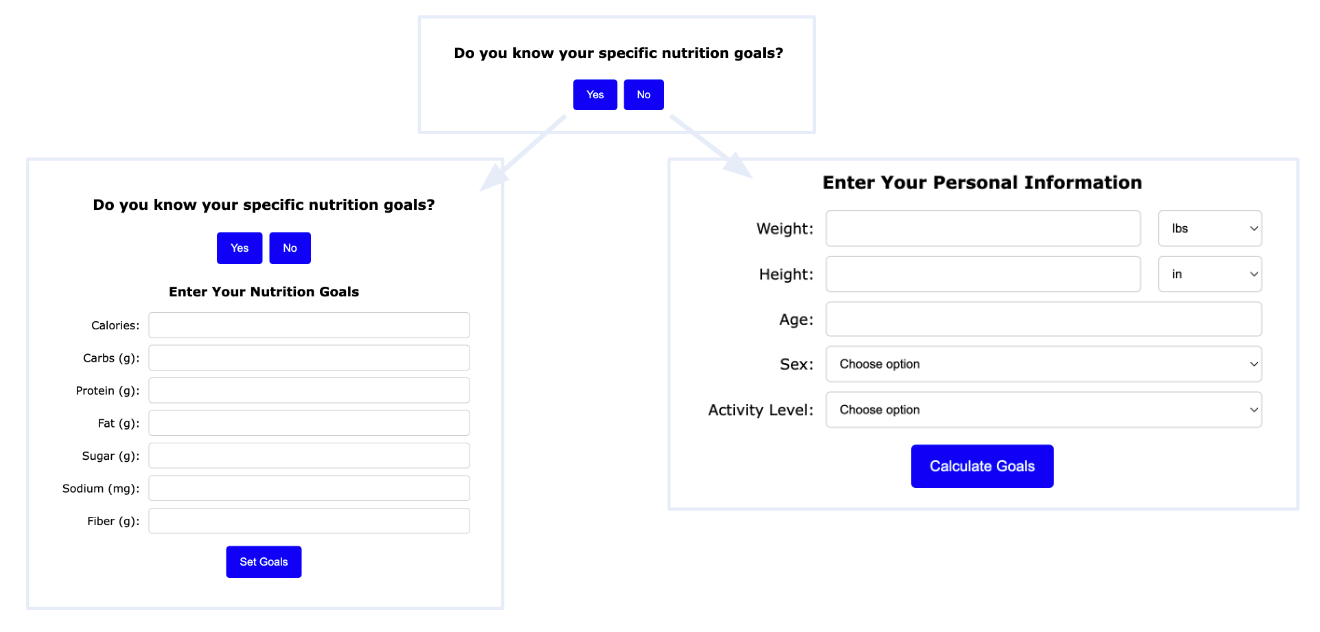}
    \caption{Meal Recommendation Prompting for User}
    \label{fig:app recs}
\end{figure}
\begin{figure}[h]
    \centering
    \includegraphics[width=\linewidth]{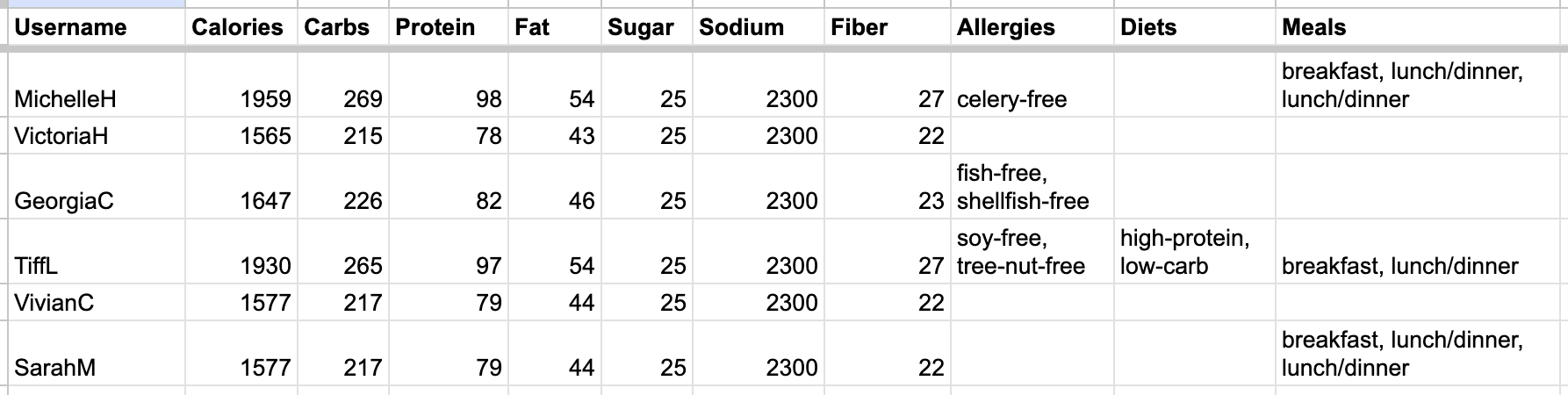}
    \caption{User information storage in Google Sheets}
    \label{fig:app gsheet}
\end{figure}
\paragraph{Server-Side Implementation} The process is managed by a Flask server, which handles requests, processes data, and returns results. The server connects to a YOLOv8 model hosted on Google Colab via ngrok for secure, real-time image processing.

\paragraph{Client-Server Implementation} The web client sends images, videos, or URLs to the server for food detection. The YOLOv8 model processes these, then sends detected items to the Edamam APIs for nutritional analysis and recommendations. The results are displayed on the web client via a user-friendly interface using Chart.js (Fig. 7).

\begin{figure}[h]
    \centering
    \includegraphics[width=\linewidth]{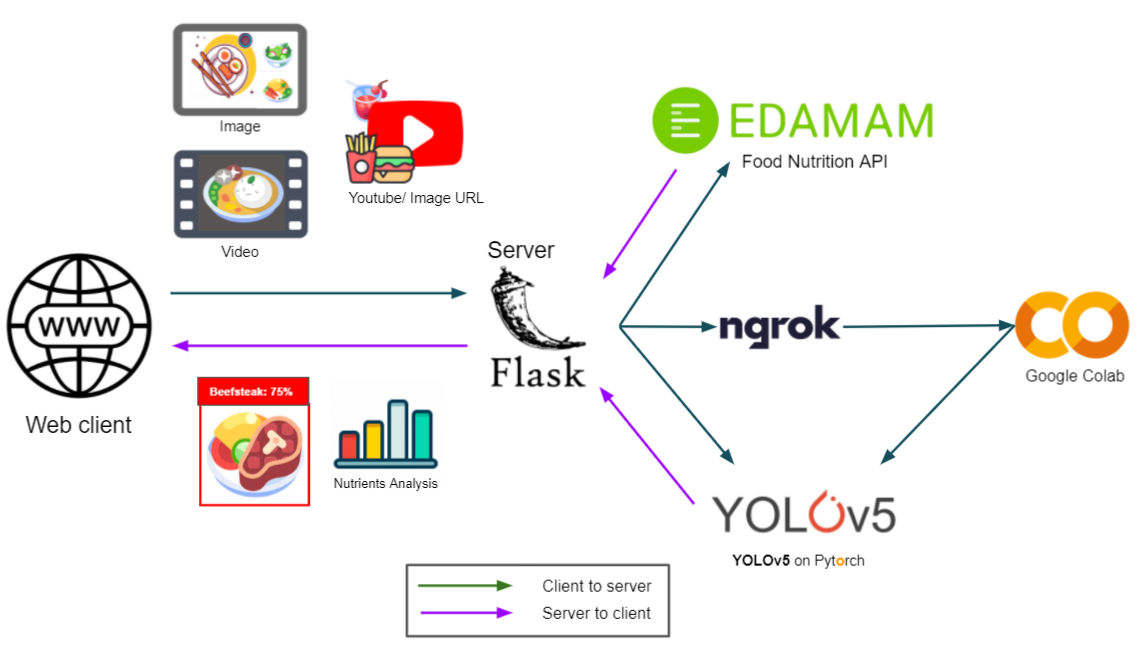}
    \caption{NutrifyAI's Client-Server and Server-Side Interaction [5]}
    \label{fig:cssss}
\end{figure}

\section{Results}
The evaluation on the Food Recognition 2022 dataset utilized metrics such as precision, recall, F1 score, and overall accuracy using the YOLOv8s that powers NutrifyAI at 0.5 IoU.

The evaluation process involved comparing model predictions to ground truth labels per image, aggregating true positives, false positives, and false negatives across all images. The model achieved an accuracy of 75.4\%, suggesting that most food items were correctly identified. Precision was 78.5\%, reflecting effective minimization of false positives, while recall was 72.8\%. The F1 score was 75.5\%, providing a balanced measure of the model's performance.

Performance varied by food class, with the model excelling in recognizing distinct food items like pomegranates and waffles, achieving near-perfect accuracy. However, it faced challenges with common or visually similar items, such as various types of apples and pears, likely due to class imbalance. Details of the performance across different food classes are illustrated in Fig. 8.

\begin{figure}[h]
    \centering
    \includegraphics[width=\linewidth]{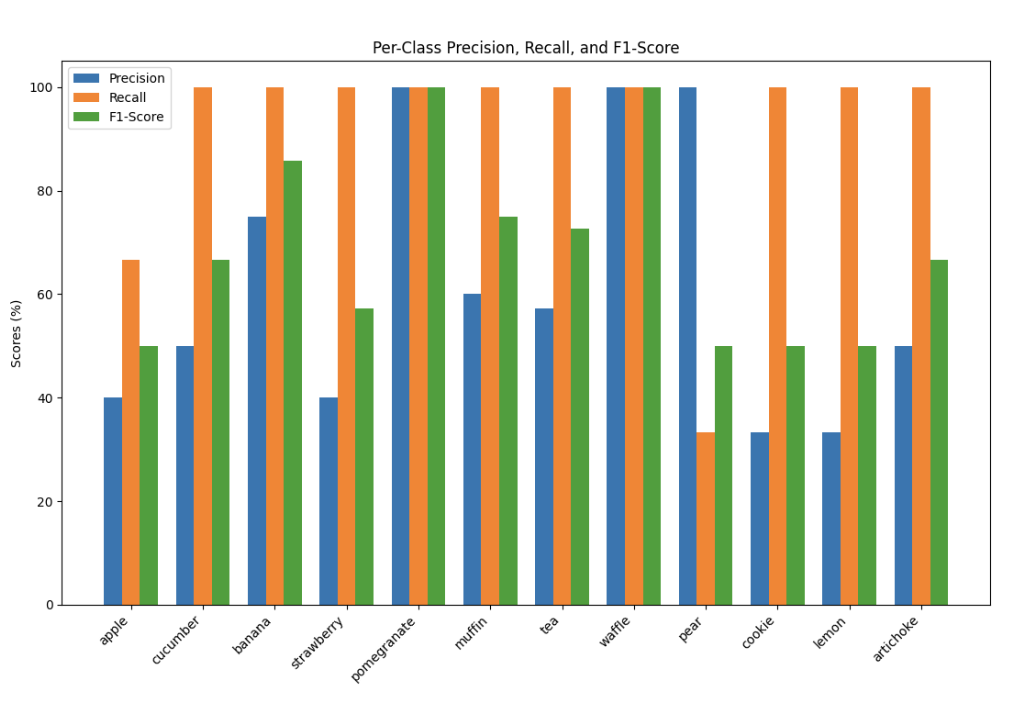}
    \caption{Precision, Recall and F1 Scores Grouped by Food Classes}
    \label{fig:stats}
\end{figure}
Additionally, the model was evaluated on detection speed. The time taken by the model to process each image and produce a ranged from 0.5s to 3.8s, with a median time of approximately 1.5 seconds, indicating that the model is efficient enough for real-time user applications.


\section{Conclusion}
NutrifyAI  demonstrates the potential of integrating computer vision models with nutritional APIs for food recognition and nutritional analysis. Despite its promising results, the system could benefit from further improvements. Future efforts should aim to diversify training datasets to enhance food item recognition and refine confidence calibration to increase reliability. Implementing user feedback mechanisms will enable the system to adapt to real-world usage, offering more personalized and accurate meal recommendations. As the field of AI-powered nutrition continues to evolve, there is significant potential for this technology to be applied in various domains, including healthcare and personalized diet planning, making the pursuit of these improvements highly worthwhile.



\begin{thebibliography}{99}

\bibitem{sharma2023} A. Sharma, ``Diet and Nutrition Apps Statistics: Exploring the Impact on Health,'' \textit{Market.us Media}, 28 Aug. 2023. \url{https://media.market.us/diet-and-nutrition-apps-statistics/}.

\bibitem{young2024} L. Young, ``Featured Blogs,'' \textit{ACSM CMS}, 2024.\url{https://www.acsm.org/blog-detail/fitness-index-blog/2024/06/12/eating-healthy-nutrition-apps}. 

\bibitem{busad2023} A. Busad \textit{et al.}, ``Computer Vision Based Food Recognition with Nutrition Analysis, 2023.\url{https://www.ijcrt.org/papers/IJCRT2301042.pdf}.

\bibitem{i3international2016} ``You Only Look Once (YOLO): The Best Model for AI-Integrated Video Analytics?, 2016.\url{https://i3international.com/resources/media/yolo-the-best-model-for-ai-integrated-video-analytics}.

\bibitem{nguyen2022} H.-L. Nguyen, ``Meal Analysis with Theseus,'' \textit{GitHub}, 21 Sept. 2022. \url{https://github.com/lannguyen0910/food-recognition}.

\bibitem{datasetninja2022} ``Food Recognition 2022 - Dataset Ninja,'' \textit{Dataset Ninja}, 2022. \url{https://datasetninja.com/food-recognition}.

\bibitem{redmon2016} J. Redmon \textit{et al.}, ``You Only Look Once: Unified, Real-Time Object Detection,'' 9 May 2016. \url{https://arxiv.org/pdf/1506.02640v5.pdf}.

\bibitem{rohini2021} B. Rohini \textit{et al.}, ``A Framework to Identify Allergen and Nutrient Content in Fruits and Packaged Food Using Deep Learning and OCR, 3 June 2021. \url{https://ieeexplore.ieee.org/document/9441800}.

\bibitem{quach2023} L.-D. Quach \textit{et al.}, ``Evaluating the Effectiveness of YOLO Models in Different Sized Object Detection and Feature-Based Classification of Small Objects, \url{https://doi.org/10.12720/jait.14.5.907-917}. 

\bibitem{reis2023} D. Reis \textit{et al.}, ``Real-Time Flying Object Detection with YOLOv8, 17 May 2023. \url{https://arxiv.org/pdf/2305.09972}.

\bibitem{sudharson2023} S. Sudharson \textit{et al.}, ``Enhanced Food Classification System Using YOLO Models for Object Detection Algorithm, 6 July 2023. \url{https://doi.org/10.1109/icccnt56998.2023.10307943}. 

\bibitem{sun2019} J. Sun \textit{et al.}, ``FoodTracker: A Real-Time Food Detection Mobile Application by Deep Convolutional Neural Networks, 13 Sept. 2019. \url{https://arxiv.org/pdf/1909.05994}.

\end{thebibliography}
\end{document}